\documentclass[11pt]{article}

\usepackage[preprint]{acl}

\usepackage{times}
\usepackage{latexsym}

\usepackage[T1]{fontenc}

\usepackage[utf8]{inputenc}

\usepackage{microtype}

\usepackage{inconsolata}

\usepackage{graphicx}

\usepackage{amsmath,amssymb}
\usepackage{booktabs}
\usepackage{multirow}
\usepackage{bbm}
\usepackage{amsthm}
\usepackage{enumitem}
\usepackage{tikz}
\usepackage{pgfplots}
\pgfplotsset{compat=1.18}

\newcommand{\method}{\texttt{SAGE}}

\title{\method: A Novelty Gate for Efficient Memory Evolution in Agentic LLMs}

\author{
Sijia Wang\thanks{These authors contributed equally to this work.}, Dhanajit Brahma\footnotemark[1], Ricardo Henao \\
Duke University \\
\texttt{\{sijia.wang, dhanajit.brahma, ricardo.henao\}@duke.edu}
}

\begin{document}
\maketitle
\begin{abstract}
Agentic LLMs must continuously decide whether newly extracted facts should be added, merged with existing memories, or ignored, yet prior work has focused more on retrieval and storage than on principled write-side control. We frame memory evolution as a novelty-detection problem and propose \method{}, a \textbf{S}pherical \textbf{A}daptive \textbf{G}ate for memory \textbf{E}volution that scores candidate facts with a von Mises-Fisher-based density estimator over memory embeddings and routes them with an adaptive threshold that tracks memory-store geometry. \method{} resolves clearly novel facts as \textsc{Add}, clearly redundant facts as \textsc{Noop}, and sends only uncertain cases to an LLM merge step, reducing expensive write-time reasoning. On LoCoMo, SAGE achieves the best average token-F1 against Mem0 on all seven open-weight backbone comparisons, while on GPT-4o-mini it reduces add-phase API cost by 3.4$\times$ and add-phase latency by 2.5$\times$ with only a small average judge-score gap. As a drop-in binary gate for A-Mem, SAGE skips roughly 16–18$\%$ of LLM calls across five models with minimal quality change on open-weight backbones. These results suggest that novelty-aware write control is a practical lever for improving both memory quality and system efficiency in long-term agentic memory. \textbf{Code available at:} \url{https://github.com/swang1024/SAGE}.
\end{abstract}

\section{Introduction}
Every memory system, from a relational database \citep{codd1970relational} to a modern LLM agent \citep{park2023generative,packer2023memgpt}, must solve three problems in sequence: decide what to \emph{write}, organize it so it can be \emph{found}, and \emph{retrieve} the right information when needed.
In agentic LLM memory, the community has invested heavily in the second and third problems -- embedding models \citep{pena2025evaluating}, vector indexes \citep{douze2025faiss, johnson2019billion}, hybrid retrieval \citep{Ma2020HybridFR, sawarkar2024blended, hsu2025dat}, knowledge graphs \citep{rasmussen2025zep}, while the first has received comparatively little principled attention.
Yet the write decision is arguably the more consequential one: a memory that is never written cannot be retrieved, and a memory that is written incorrectly (duplicated, merged with an unrelated fact, or prematurely deleted) will degrade downstream queries that touch it.
How difficult this write decision is depends on the memory paradigm.

While standard Retrieval-Augmented Generation (RAG) writes are nearly decision-free: segment, embed, append \citep{karpukhin2020dense},
long-term agentic systems cannot afford this luxury. An agent interacting over weeks or months must track an evolving state--changing preferences, shifting goals, and corrected facts. This forces agentic memory systems to confront the dilemma of semantic CRUD~\citep{lyu2025crud, lee2024human}: they must edit their own knowledge base in natural language, continuously deciding whether to add, update, consolidate, or discard information rather than simply accumulating it.
Current systems delegate this decision to an LLM: Mem0 issues a tool call that jointly routes and rewrites each batch of extracted facts \citep{mem0}; A-Mem adds further calls for note construction and neighbor evolution \citep{xu2025mem}.
These designs produce adaptive memory stores, but make the write path the dominant source of cost.
We argue that the missing alternative is a \emph{novelty gate}: a cheap, closed-form test that routes clearly new facts to \textsc{Add}, clearly redundant facts to \textsc{Noop}, and only ambiguous cases to an LLM merge call.

The paper makes three contributions: {\em i)} It frames memory evolution in agentic LLMs as a novelty-detection problem, clarifying why write-side control is the lever that affects both memory quality and system efficiency. {\em ii)} It proposes \method{} (\textbf{S}pherical \textbf{A}daptive \textbf{G}ate for memory \textbf{E}volution), a theoretically grounded novelty gate whose score is computed using vMF density estimation, together with an adaptive threshold that tracks the evolving geometry of the memory store. {\em iii)} It provides evidence across two settings: as a full system, \method{} wins 7/7 open-weight backbones on token-$F_1$ against Mem0 while cutting add-phase API cost $3.4\times$ on GPT-4o-mini; as a drop-in \textsc{Noop} gate on A-Mem, it skips 16--18\% of write LLM calls across five models with $\leq$0.5\% token-$F_1$ change.

\section{Related Work}

\noindent\textbf{Memory for Agentic LLMs.}
Long-term memory has become a central topic in LLM-agent research because raw context extension does not reliably solve multi-session reasoning \citep{zhang2024survey,maharana2024evaluating}. Prior work falls into three broad categories. \emph{Retrieval and compression} methods reduce long histories to retrievable summaries: MemoryBank \citep{zhong2024memorybank} applies Ebbinghaus-inspired forgetting, ReadAgent \citep{readagent} compresses conversations into gist memories, and Generative Agents \citep{park2023generative} consolidate observations through periodic LLM-driven reflection. \emph{Structured and hierarchical} approaches impose richer organization: Zep \citep{rasmussen2025zep} and Mem$0_g$ \citep{mem0} maintain temporal or entity-relation knowledge graphs, while MemGPT \citep{packer2023memgpt} introduces OS-style paging between working memory and an external store. Finally, \emph{learned representations} such as MEM1 \citep{mem12025} train a compact internal state via end-to-end RL. Across all three categories, write policies remain either fixed (append-only, forgetting curves, heuristic eviction) or fully delegated to per-fact LLM judgment; efficient write-side control of memory evolution remains an open problem.

\begin{figure*}[t]
\centering
\includegraphics[width=0.9\textwidth]{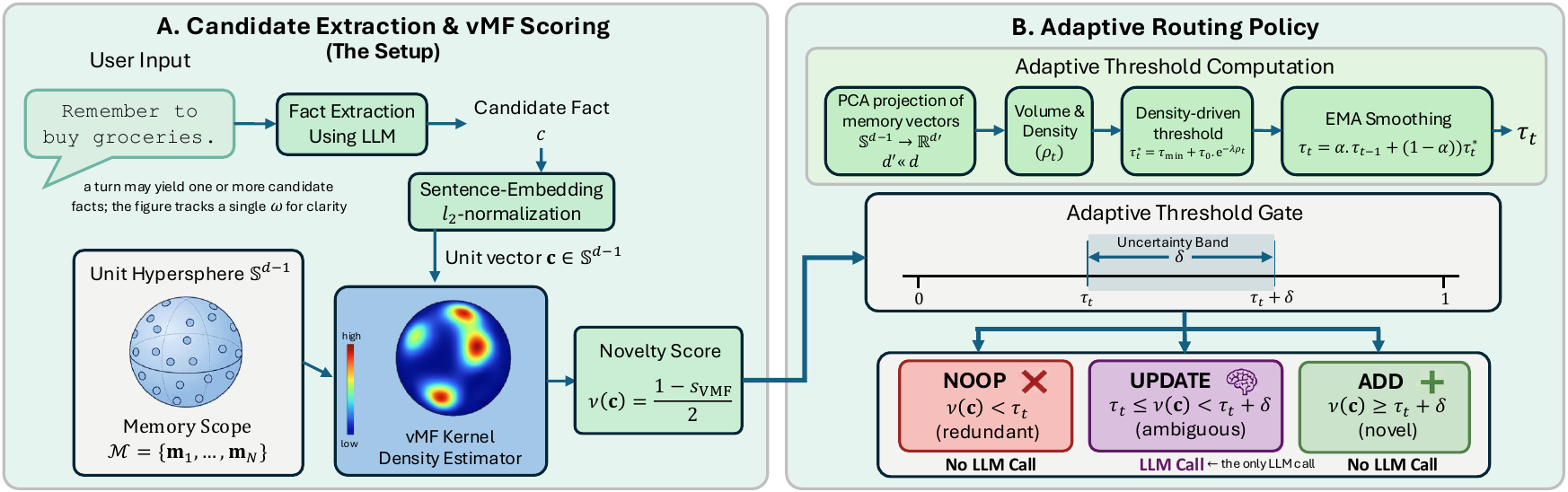}
\vspace{-0.5em}
\caption{Overview of Memory Evolution problem and our proposed approach \method{}.
}
\vspace{-1.5em}
\label{fig:overview}
\end{figure*}
 
\noindent\textbf{Memory Evolution.}
Recent agentic memory systems treat memory as an editable structure rather than an append-only log. Mem0 \citep{mem0} extracts salient facts and uses an LLM-mediated controller to choose among \textsc{Add}, \textsc{Update}, \textsc{Delete}, and \textsc{Noop}. A-Mem \citep{xu2025mem} extends this to full memory evolution, constructing structured notes with contextual descriptions and rewriting linked neighbors as new evidence arrives. A newer line replaces prompted write control with reinforcement learning: Memory-R1 \citep{yan2025memory} trains a dedicated memory manager via PPO/GRPO, with reward derived from downstream QA performance, and Mem-$\alpha$ \citep{wang2025mem} similarly uses RL to optimize memory construction across core, episodic, and semantic stores, demonstrating strong length generalization. Overall, prior work shows that write-side memory control is essential, but existing approaches sit at two costly extremes: repeated LLM-based deliberation at inference time or rollout-intensive RL optimization at training time. Our work explores a third point in the design space, treating memory evolution as a novelty-aware control problem in which the system first estimates whether an incoming fact is sufficiently new to justify memory editing. This framing yields a lightweight, geometry-aligned controller that preserves the benefits of adaptive memory evolution while avoiding both the inference overhead of pure LLM routing and the training overhead of RL-based policy learning. 

\section{Methodology}\label{sec:methodology}
An agentic LLM memory system maintains a persistent store of facts and observations across conversation sessions. In each user interaction, it extracts candidate facts, such as preferences, goals, or contextual details, from the current turn. 
For each candidate, the system makes a \(\emph{write\text{-}side}\) decision among three actions: 
\textsc{Add}, which stores the fact as a new memory; \textsc{Update}, which merges the fact with an existing memory that it refines, corrects, or supersedes; and \textsc{Noop}, which ignores the fact because the information is already covered by the current memory store. We call the component that makes this decision the \(\emph{routing\ controller}\). Figure~\ref{fig:overview} summarizes this workflow and shows where the novelty score-based gating operates relative to candidate fact extraction, novelty scoring, and update-time reasoning.
In this section, we formalize write-side memory control as a novelty-detection problem and introduce \(\method{}\) (Spherical Adaptive Gate for memory Evolution) as the routing controller. We first define the problem, then motivate the von Mises-Fisher (vMF) distribution as the foundation of a kernel density estimator for scoring how novel each candidate fact is relative to the current memory store and route it to \textsc{Add}, \textsc{Update}, or \textsc{Noop} via an adaptive threshold.

\subsection{Problem Definition}
We begin by defining the system components before formalizing the decision problem. A \emph{stored memory item} is a candidate fact previously extracted from a user interaction and committed to persistent storage ({\em e.g.}, ``the user prefers morning meetings''). Each memory item is embedded by a sentence embedding model~\citep{sentence-bert} and $\ell_2$-normalized onto the unit hypersphere $\mathbb{S}^{d-1} = \{\mathbf{z} \in \mathbb{R}^d : \|\mathbf{z}\|_2 = 1\}$. 
The current memory scope is therefore a set of unit-norm embedding vectors $\mathcal{M} = \{\mathbf{m}_1,\ldots,\mathbf{m}_N\}$, where $\mathbf{m}_i \in \mathbb{S}^{d-1}$.
In practice, this scope consists of the stored memory items paired with their embedding vectors: the downstream memory writing and rewriting operate on the associated memory items, as in prior works such as Mem0~\cite{mem0} and A-Mem~\cite{xu2025mem}, while the embedding vectors are used during routing or retrieval.
During each \emph{interaction} (a conversation turn or session), the system extracts one or more candidate facts by making an LLM call, again following the fact-extraction stage used in systems such as Mem0 and A-Mem.
Let \textit{c} denote a candidate fact and $\mathbf{c} \in \mathbb{S}^{d-1}$ its normalized embedding.
Then the routing controller must decide which decision to make given a candidate fact $c$.

\subsection{From Memory Evolution to Novelty Detection}
Routing is difficult because different mistakes have different costs: an overly conservative controller discards new information; an overly permissive one accumulates near-duplicates that degrade retrieval; and an unreliable one may conflate related but distinct facts ({\em e.g.}, merging ``flight departs at 8\,am'' with ``meeting starts at 8\,am''), corrupting accurate records.
Mem0~\citep{mem0} invokes an LLM controller on every batch of candidate facts regardless of novelty; 
A-Mem~\citep{xu2025mem} adds further LLM calls for note construction and for rewriting nearby stored memories to keep related notes consistent. 
In both, routing cost scales with {\em all} candidate facts.

We therefore introduce a novelty score as a first routing stage before any update-time LLM call. The goal is to separate candidates that are likely new from those that are likely redundant, and to send only the remaining uncertain cases to the LLM update step. 
Here, an uncertain case is one whose score does not strongly favor either \textsc{Add} or \textsc{Noop}. 
This gate reduces write-time cost by reserving LLM-based updates for those cases rather than for every candidate.
In our experiments, this decision stage reduces LLM calls by $60$--$90\%$ compared to Mem0 on seven of the eight backbones.
To our knowledge, existing memory-evolution systems do not include this kind of explicit routing gate; however, this is largely because prior work prioritized memory quality and adaptivity over minimizing controller cost at write time. The next section specifies the gate itself.

The embedding geometry also suggests how to build this gate. Sentence-embedding memory systems operate on $\ell_2$-normalized vectors compared by cosine similarity~\citep{sentence-bert,karpukhin2020dense}, which for unit vectors is simply their inner product, so semantic comparison is driven by direction rather than by magnitude.

Novelty in this setting should not depend only on the closest stored memory but also on how much support the surrounding memories provide. For example, two candidates can have the same cosine similarity to a memory item yet differ in novelty to the memory scope: one may lie in a region already populated by several similar memories, while the other lies near a more isolated memory. The first candidate is less novel because it is better supported by the existing memory set. 

These observations suggest that the novelty score-based inexpensive routing rule should: ($i$)~be computationally cheap so that many candidates can be resolved without an LLM call, ($ii$)~operate in the same inner-product geometry as retrieval, and ($iii$)~account for how densely populated the nearby stored memories are to estimate if the candidate is redundant. 
A natural way to capture this support is kernel density estimation (KDE), which scores a point by placing a local kernel around each stored memory and summing their contributions. 
Because the embeddings are unit-norm directional vectors and retrieval depends on angular similarity, we use a kernel that depends only on direction. The von Mises--Fisher (vMF) distribution~\citep{mardia1999directional,banerjee2005clustering} is a standard model for directional data on \(\mathbb{S}^{d-1}\), so it is an appropriate kernel for spherical KDE.
A vMF with mean direction \(\boldsymbol{\mu} \in \mathbb{S}^{d-1}\) and concentration \(\kappa > 0\) has density
$f(\mathbf{c} \mid \boldsymbol{\mu}, \kappa) = C_d(\kappa)\exp(\kappa\,\boldsymbol{\mu}^{\top}\mathbf{c}),
$
where \(C_d(\kappa)\) is a normalizing constant depending only on \(d\) and \(\kappa\). In our KDE, this density serves as the kernel centered at each stored memory vector. Since it depends only on inner product \(\boldsymbol{\mu}^{\top}\mathbf{c}\), it is well suited to modeling local support on the hypersphere.

\subsection{\method{}: Spherical Adaptive Gate for Memory Evolution} \label{sec:methodsage}
Given a candidate embedding $\mathbf{c} \in \mathbb{S}^{d-1}$ and the current memory scope $\mathcal{M}$, the goal is to obtain a scalar novelty score that quantifies how well the direction of $\mathbf{c}$ is explained by the stored memory embeddings. We define this score via a kernel density estimate on the hypersphere.

To estimate the density that $\mathcal{M}$ induces at $\mathbf{c}$, we center a vMF-inspired kernel at each stored memory vector and average across memories.
We therefore work with the kernel
$
K_\kappa(\mathbf{c},m_i)=\exp(\kappa\, m_i^\top \mathbf{c}),
$
which retains the angular structure of the vMF distribution while avoiding unnecessary terms.
Averaging over the memory scope gives
$
\hat{S}(\mathbf{c}\mid\mathcal{M})=\frac{1}{N}\sum_{i=1}^{N}K_\kappa(\mathbf{c},m_i).
$
This average is well defined for $N \ge 1$, since it is a finite sum of positive, bounded terms. When $N=0$ (i.e., the memory scope is empty), the controller directly emits \textsc{Add} without computing a score.
Taking the logarithm and dividing by $\kappa$ keeps the result on the cosine-similarity scale; Appendix~\ref{app:vmf-score-bound} shows that for \(N\geq 1\), the resulting score lies in \([-1,1]\). This yields
$
s_{\mathrm{vMF}}(\mathbf{c}\mid\mathcal{M})=\frac{1}{\kappa}\log\hat{S}(\mathbf{c}\mid\mathcal{M}).
$
Structurally, $s_{\mathrm{vMF}}$ is the log-mean-exp of the cosine similarities $\{m_i^\top \mathbf{c}\}$, scaled by $\frac{1}{\kappa}$. It therefore produces a single scalar that summarizes how much collective angular support the entire memory scope provides for $\mathbf{c}$. 

Unlike raw cosine similarity, which compares $\mathbf{c}$ to one memory at a time, $s_{\mathrm{vMF}}$ aggregates contributions from all stored memories. Consequently, a candidate that has a high cosine similarity to a single isolated memory can still receive a different $s_{\mathrm{vMF}}$ score than a candidate with the same cosine similarity score in a densely populated region of supporting memories. 
$
\nu(\mathbf{c})=\frac{1-s_{\mathrm{vMF}}(\mathbf{c}\mid \mathcal{M})}{2}$.
This affine transformation does not change the ranking of candidates; it is used only so that larger values mean ``more novel,'' which simplifies the interpretation of the adaptive threshold and margin defined in Section~\ref{sec:adaptive-routing}.

The concentration parameter $\kappa$ is not fixed a priori but is estimated from the current memory scope so that the gate adapts to the geometry of the stored embeddings. We compute the mean resultant length
$
\bar{R} = \left\|\frac{1}{N}\sum_{i=1}^{N}\mathbf{m}_i\right\|_2,
$
which measures the concentration of the memory vectors around their mean direction ($\bar{R} \approx 1$
 when the vectors are tightly concentrated, $\bar{R} \approx 0$ when they are diffusely distributed). Following~\citet{banerjee2005clustering}, we estimate $\kappa$ via the approximation
$
\hat{\kappa} \approx \frac{\bar{R}(d-\bar{R}^2)}{1-\bar{R}^2}
$
which ensures that $\hat{\kappa}$ adapts to how spread out the stored memories are. When memories are densely stored, $\hat{\kappa}$ is large, and the score is more sensitive to small directional differences; when scattered, $\hat{\kappa}$ is small and each kernel covers a wider region.

This is the {\em key advantage} over a cosine-similarity-based threshold: as $\mathcal{M}$ changes, $\hat{\kappa}$ adapts automatically, so the effective influence of each stored memory reflects the current density of the store rather than remaining fixed.

\begin{table*}[t]
\centering
\small
\setlength{\tabcolsep}{3.pt}
\renewcommand{\arraystretch}{1.10}
\resizebox{\textwidth}{!}{%
\begin{tabular}{cc|ccc|ccc|ccc|ccc|ccc}
\toprule
\multirow{2}{*}{\textbf{Model}} & \multirow{2}{*}{\textbf{Method}} & \multicolumn{3}{c|}{\textbf{Single-hop}} & \multicolumn{3}{c|}{\textbf{Multi-hop}} & \multicolumn{3}{c|}{\textbf{Temporal}} & \multicolumn{3}{c|}{\textbf{Open-domain}} & \multicolumn{3}{c}{\textbf{Average}} \\
& & $F_1\uparrow$ & $B_1\uparrow$ & $J\uparrow$ & $F_1\uparrow$ & $B_1\uparrow$ & $J\uparrow$ & $F_1\uparrow$ & $B_1\uparrow$ & $J\uparrow$ & $F_1\uparrow$ & $B_1\uparrow$ & $J\uparrow$ & $F_1\uparrow$ & $B_1\uparrow$ & $J\uparrow$ \\
\midrule
\multirow{3}{*}{\shortstack[c]{DeepSeek-R1\\1.5B}} & \method{} & \textbf{8.81} & \textbf{6.51} & \textbf{93.62} & \textbf{8.07} & \textbf{4.91} & \textbf{87.54} & \textbf{9.80} & \textbf{7.36} & 87.50 & \textbf{10.67} & \textbf{5.77} & \textbf{94.05} & \textbf{9.34} & \textbf{6.14} & \textbf{90.68} \\
 & Mem0 & 6.31 & 4.99 & 89.72 & 7.36 & 4.42 & 85.36 & 7.58 & 4.83 & \textbf{92.71} & 8.21 & 4.21 & 93.10 & 7.36 & 4.61 & 90.22 \\
 & $\text{Mem0}\textsuperscript{\textit{g}}$ & 6.64 & 5.22 & 89.72 & 6.72 & 4.21 & 83.49 & 8.49 & 6.01 & 89.58 & 8.00 & 4.40 & 91.44 & 7.46 & 4.96 & 88.56 \\
\midrule
\multirow{3}{*}{\shortstack[c]{DeepSeek-R1\\7B}} & \method{} & 14.98 & 11.26 & \textbf{98.58} & \textbf{14.72} & \textbf{8.80} & \textbf{88.16} & \textbf{9.64} & 7.22 & 87.50 & \textbf{14.76} & 8.76 & \textbf{94.77} & \textbf{13.52} & 9.01 & 92.25 \\
 & Mem0 & \textbf{15.41} & \textbf{11.57} & 96.10 & 13.54 & 8.27 & 81.62 & 9.21 & \textbf{7.27} & 92.71 & 14.56 & \textbf{9.04} & 94.17 & 13.18 & \textbf{9.04} & 91.15 \\
 & $\text{Mem0}\textsuperscript{\textit{g}}$ & 13.07 & 9.87 & 93.26 & 13.73 & 8.62 & 85.36 & 9.53 & 6.94 & \textbf{96.88} & 14.42 & 8.76 & 94.05 & 12.69 & 8.55 & \textbf{92.39} \\
\midrule
\multirow{3}{*}{\shortstack[c]{Llama-3.2\\1B}} & \method{} & \textbf{5.64} & \textbf{3.68} & \textbf{84.75} & \textbf{4.67} & \textbf{2.02} & \textbf{79.44} & \textbf{5.57} & \textbf{2.57} & \textbf{86.46} & \textbf{6.52} & \textbf{2.49} & \textbf{90.01} & \textbf{5.60} & \textbf{2.69} & \textbf{85.16} \\
 & Mem0 & 2.70 & 1.85 & 59.57 & 2.43 & 0.87 & 57.94 & 4.48 & 2.00 & 55.21 & 3.15 & 1.10 & 59.22 & 3.19 & 1.46 & 57.98 \\
 & $\text{Mem0}\textsuperscript{\textit{g}}$ & 2.97 & 2.05 & 58.87 & 2.21 & 0.91 & 55.45 & 4.34 & 1.93 & 65.62 & 3.05 & 1.07 & 60.88 & 3.14 & 1.49 & 60.20 \\
\midrule
\multirow{3}{*}{\shortstack[c]{Llama-3.2\\3B}} & \method{} & \textbf{5.03} & 2.42 & \textbf{82.62} & \textbf{4.98} & 1.68 & \textbf{76.01} & \textbf{5.22} & \textbf{2.34} & 75.00 & \textbf{6.10} & \textbf{1.94} & \textbf{75.86} & \textbf{5.33} & \textbf{2.09} & \textbf{77.37} \\
 & Mem0 & 4.53 & 2.40 & 73.05 & 4.74 & \textbf{1.70} & 72.59 & 4.49 & 2.03 & 72.92 & 5.88 & 1.91 & 72.18 & 4.91 & 2.01 & 72.68 \\
 & $\text{Mem0}\textsuperscript{\textit{g}}$ & 4.74 & \textbf{2.44} & 74.47 & 4.63 & 1.66 & 68.54 & 4.97 & 2.33 & \textbf{81.25} & 5.56 & 1.80 & 72.65 & 4.97 & 2.06 & 74.23 \\
\midrule
\multirow{3}{*}{\shortstack[c]{Qwen2.5\\1.5B}} & \method{} & 5.88 & 4.60 & \textbf{69.50} & \textbf{25.30} & \textbf{22.89} & 57.94 & 11.25 & 10.18 & 67.71 & \textbf{8.04} & \textbf{6.81} & 78.60 & \textbf{12.62} & \textbf{11.12} & 68.44 \\
 & Mem0 & \textbf{6.19} & 4.68 & 68.44 & 24.55 & 21.08 & \textbf{65.73} & \textbf{12.34} & \textbf{10.77} & \textbf{68.75} & 7.28 & 5.72 & \textbf{79.19} & 12.59 & 10.56 & \textbf{70.53} \\
 & $\text{Mem0}\textsuperscript{\textit{g}}$ & 5.63 & \textbf{4.90} & 63.83 & 25.21 & 21.88 & 62.93 & 9.58 & 9.49 & 60.42 & 7.74 & 6.12 & 78.12 & 12.04 & 10.60 & 66.32 \\
\midrule
\multirow{3}{*}{\shortstack[c]{Qwen2.5\\3B}} & \method{} & \textbf{28.27} & \textbf{18.63} & \textbf{82.98} & 31.00 & 27.66 & 78.19 & \textbf{12.05} & \textbf{10.07} & 67.71 & \textbf{35.69} & \textbf{30.12} & 90.61 & \textbf{26.75} & \textbf{21.62} & 79.87 \\
 & Mem0 & 26.70 & 18.48 & 82.27 & 31.63 & 27.88 & 77.57 & 10.58 & 9.38 & 64.58 & 33.25 & 28.76 & \textbf{90.84} & 25.54 & 21.12 & 78.82 \\
 & $\text{Mem0}\textsuperscript{\textit{g}}$ & 25.94 & 17.82 & \textbf{82.98} & \textbf{32.28} & \textbf{28.95} & \textbf{78.82} & 11.75 & 9.43 & \textbf{71.88} & 31.98 & 26.94 & 88.35 & 25.49 & 20.78 & \textbf{80.51} \\
\midrule
\multirow{3}{*}{\shortstack[c]{Qwen2.5\\7B}} & \method{} & 26.58 & \textbf{17.89} & 83.69 & \textbf{28.36} & \textbf{24.22} & \textbf{73.52} & \textbf{12.30} & \textbf{10.83} & 76.04 & \textbf{36.30} & \textbf{31.22} & \textbf{92.75} & \textbf{25.88} & \textbf{21.04} & \textbf{81.50} \\
 & Mem0 & 25.97 & 17.19 & \textbf{85.82} & 25.52 & 21.44 & 65.11 & 11.98 & 9.14 & \textbf{77.08} & 32.95 & 28.65 & 90.84 & 24.10 & 19.10 & 79.71 \\
 & $\text{Mem0}\textsuperscript{\textit{g}}$ & \textbf{26.75} & 17.36 & 84.40 & 23.59 & 20.35 & 62.31 & 11.87 & 8.70 & 73.96 & 31.86 & 27.76 & 87.75 & 23.52 & 18.54 & 77.10 \\
 \midrule
 \multirow{2}{*}{\shortstack[c]{GPT-4o\\mini}} & \method{} & 32.46 & 19.81 & 53.90 & \textbf{49.61} & 40.91 & \textbf{56.07} & 19.87 & 14.73 & 35.42 & \textbf{44.79} & 36.65 & \textbf{63.26} & 36.69 & 28.02 & 52.16 \\
 & Mem0 & \textbf{34.10} & \textbf{20.86} & \textbf{56.03} & 48.75 & \textbf{41.05} & 52.34 & \textbf{20.77} & \textbf{15.51} & \textbf{42.71} & 44.02 & \textbf{36.71} & 62.90 & \textbf{36.91} & \textbf{28.53} & \textbf{53.50} \\
 
 \bottomrule
\end{tabular}
}
\vspace{-0.3em}
\caption{Detailed per-configuration comparison across \method{}, Mem0, and $\text{Mem0}\textsuperscript{\textit{g}}$. 
Metrics are mean token-$F_1$, BLEU-1 ($B_1$), and LLM-as-a-Judge ($J$).
}
\vspace{-1em}
\label{tab:overall-headtohead}
\end{table*}

\subsection{Adaptive Routing Rule} 
\label{sec:adaptive-routing}

Since $\nu(\mathbf{x})$ is defined relative to current memory scope, the same raw novelty score can mean different things in sparse and dense stores; we therefore adapt the routing threshold to a simple proxy for how tightly the current memories are packed.

We use a proxy $\rho_t$ to quantify the density of the current memory scope, and we provide the details in Appendix~\ref{app:proxyrho}.
A larger \(\rho_t\) means that many memories occupy a relatively small region of the informative subspace. In such cases, novelty scores are typically pushed downward because candidates are more likely to land near already crowded regions, so the gate should become more permissive. This motivates the monotone decay $
\tau_t^\star=\tau_{\min}+\tau_0 e^{-\lambda \rho_t},$
where \(\tau_0\) is the base threshold, \(\tau_{\min}\) is a floor, and \(\lambda\) controls how quickly the threshold relaxes as the scope becomes denser.
$\tau_t^\star$ decays monotonically toward $\tau_{\min}$ as density increases, and we smooth the threshold via an exponential moving average (EMA) to prevent abrupt shifts when a single turn adds several memories:
\begin{equation}
\tau_t =
\begin{cases}
\tau_t^\star, & t = 1,\\
\alpha\, \tau_{t-1} + (1-\alpha)\,\tau_t^\star, & t > 1,
\end{cases}
\end{equation}
where $\alpha \in [0,1)$ is the EMA momentum.

Using $\tau_t$ together with a margin $\delta$, the routing rule is
{
\begin{equation*}
\text{route}(\mathbf{c}) =
\begin{cases}
\textsc{Add},    & N = 0,\\
\textsc{Add},    & \nu(\mathbf{c}) \ge \tau_t + \delta,\\
\textsc{Update}, & \tau_t \le \nu(\mathbf{c}) < \tau_t + \delta,\\
\textsc{Noop},   & \nu(\mathbf{c}) < \tau_t.
\end{cases}
\end{equation*}
}
The margin $\delta$ defines an uncertainty band around the threshold, following the principle of classification with a reject option~\cite{chowadaptthresh}. Candidates above the band are routed to \textsc{Add} and those below to \textsc{Noop}, both without an LLM call. 
Only candidates within the band, i.e., genuinely ambiguous cases whether the candidate is novel enough to the scope, trigger an LLM \textsc{Update} call. 
Appendix \ref{app:illustrateadapt} provides a detailed visual trace of this process, illustrating how the adaptive threshold $\tau_t$ decays over time to accommodate increasing memory density, and how the uncertainty margin $\delta$ cleanly separates these three routing decisions.

\begin{table*}[t]
\centering
\small
\setlength{\tabcolsep}{6pt}
\renewcommand{\arraystretch}{1.25}
\begin{tabular}{l|ccc|ccc|ccc}
\toprule
Category & \multicolumn{3}{c|}{$B_1$$\uparrow$} & \multicolumn{3}{c|}{$F_1\uparrow$} & \multicolumn{3}{c}{$J$$\uparrow$} \\
 & \method{} & Mem0 & $\text{Mem0}\textsuperscript{\textit{g}}$ & \method{} & Mem0 & $\text{Mem0}\textsuperscript{\textit{g}}$ & \method{} & Mem0 & $\text{Mem0}\textsuperscript{\textit{g}}$ \\
\midrule
 Single-hop & \textbf{9.28} & 8.74 & 8.52 & \textbf{13.60} & 12.54 & 12.25 & \textbf{85.11} & 79.28 & 78.22 \\
 Multi-hop & \textbf{13.17} & 12.24 & 12.37 & \textbf{16.73} & 15.68 & 15.48 & \textbf{77.26} & 72.27 & 70.99 \\
 Temporal & \textbf{7.22} & 6.49 & 6.40 & \textbf{9.40} & 8.67 & 8.65 & \textbf{78.27} & 74.85 & 77.08 \\
 Open-domain & \textbf{12.44} & 11.34 & 10.98 & \textbf{16.87} & 15.04 & 14.66 & \textbf{88.09} & 82.79 & 81.89 \\
\bottomrule
\end{tabular}
\vspace{-0.3em}
\caption{Macro category averages across seven open-weight models. 
}
\vspace{-1.5em}
\label{tab:category-headtohead}
\end{table*}

\subsection{Extending the Gate to Other Memory Systems}
\label{sec:leakfree}
Any memory system that processes every incoming candidate \textit{c} through its full write path incurs an LLM call even when \textit{c} is clearly redundant given the current memory scope $\mathcal{M}$. A natural question is whether the vMF novelty score from Section~\ref{sec:methodsage} can serve as a lightweight pre-filter that sits upstream of any existing memory system and filters out candidates that the store already covers well.

Thus, we define a portable binary gate that can sit upstream of any existing memory system. Unlike the three-way adaptive rule in Section~\ref{sec:adaptive-routing}, this gate uses a single fixed threshold $\tau_{\text{noop}}$ and makes one decision:
\begin{equation*}
\text{route}(\mathbf{c}) =
\begin{cases}
\textsc{Noop}, & s_{\text{vMF}}(\mathbf{c}\mid\mathcal{M}) > \tau_{\text{noop}},\\
\textsc{Pass},  & \text{otherwise},
\end{cases}
\end{equation*}
where \textsc{Pass} forwards the candidate to the host system unchanged. When $s_{\text{vMF}}$ exceeds $\tau_{\text{noop}}$, the candidate \textit{c} is sufficiently explained by existing memories and is dropped; otherwise, the host system (A-Mem, Mem0, or any comparable framework) processes it with its own evolution logic fully intact. The gate is {\em non-invasive}: it exposes a single tunable knob $\tau_{\text{noop}}$ and requires no modification to the host's internals. Moreover, $\tau_{\text{noop}}$ can be set {\em without access to the target benchmark} via the calibration procedure described in Appendix~\ref{app:leakfree-calib}.

\vspace{-0.25em}
\section{Experiments}
\vspace{-0.5em}
\paragraph{Experimental Setting.} We focus on long-term conversational memory, using LoCoMo as the main benchmark protocol since it directly evaluates whether a system can answer questions from extended, multi-session dialogue histories \citep{maharana2024evaluating} (see Appendix~\ref{app:locomo-dataset} for dataset details). Following prior work, we consider single-hop, multi-hop, temporal, and open-domain questions, and we evaluate with BLEU-1 ($B_1$), token-$F_1$ ($F_1$), and LLM-as-a-Judge \citep{xu2025mem} ($J$). 
Our main experimental comparison uses seven backbone configurations for which scored \method{}/Mem0/$\text{Mem0}\textsuperscript{\textit{g}}$ are available.
We use Llama-3.1-8b as the LLM Judge model.

Prior memory papers inform the broader baseline landscape.
A-Mem compares against LoCoMo, ReadAgent, MemoryBank, and MemGPT, and reports strong gains together with write-time efficiency from selective top-$k$ retrieval \citep{xu2025mem}. Mem0 emphasizes scalable memory extraction and update-time routing over salient facts rather than full-context prompting \citep{mem0}. 
We do not re-run these comparisons; instead, we focus on the question they leave open: can a principled novelty gate replace the controller LLM in the write path?
We also test \method{} on a frontier-class backbone (GPT-4o-mini) in Section~\ref{sec:leakfree-exp} to assess whether the gate's advantages persist when the underlying LLM is strong enough to route accurately on its own.
\method{} uses the following hyperparameters for the novelty-routing gate (Section~\ref{sec:adaptive-routing}). We set the PCA projection dimension $d' = 16$. 
For the adaptive threshold $\tau_t$ update, the base threshold parameter is set to $\tau_0 = 0.25$, the minimum threshold floor is set to $\tau_{\min} = 0.025$, and the density decay coefficient is set to $\lambda = 2.0$. The temporal EMA smoothing coefficient is set to $\alpha = 0.9$. The uncertainty band is defined by $\delta = 0.025$. 
Appendix~\ref{app:hyperparameter-details} describes the selection procedure.
\vspace{-0.4em}
\subsection{Results}
Table~\ref{tab:overall-headtohead} compares \method{} against Mem0 and $\text{Mem0}\textsuperscript{\textit{g}}$ across seven backbone-matched triads. The clearest result is consistency on $F_1$: \method{} ranks first on the overall average for all seven backbones. It also achieves the best overall $B_1$ in six of seven triads, with DeepSeek-R1-7b as the only exception, where Mem0 is marginally higher (9.04 vs.\ 9.01). $J$ scores are more mixed, but still favorable to \method{} overall: it attains the best average $J$ score in four triads, exceeds Mem0 in six of seven, and exceeds $\text{Mem0}\textsuperscript{\textit{g}}$ in five of seven. Among the \method{} variants, Qwen2.5-3b is strongest on $F_1$ and $B_1$, while DeepSeek-R1-7b gives the highest average $J$ score.

Table~\ref{tab:category-headtohead} shows the same pattern after averaging by question type. \method{} is the only system that ranks first on $B_1$, $F_1$, and $J$ in all four categories. The largest $F_1$ gain over Mem0 appears on open-domain questions (+1.83), and the largest $J$ gain appears on single-hop questions (+5.83). Multi-hop gains are also steady, with \method{} reaching 16.73 $F_1$ and 77.26 $J$ versus 15.68 and 72.27 for Mem0, which suggests that better write-side separation of related facts helps later composition rather than only surface overlap. Temporal questions remain the tightest comparison, but \method{} still leads there on all three metrics.

\begin{table*}[t]
\centering
\small
\setlength{\tabcolsep}{6pt}
\renewcommand{\arraystretch}{0.9}
\begin{tabular}{cc ccccccc}
\toprule
\multirow{2}{*}{\textbf{Model}} & \multirow{2}{*}{\textbf{System}} & \multicolumn{2}{c}{\textbf{Decision-Stage LLM}} & \textbf{Extraction} & \textbf{Total write} & \multirow{2}{*}{\textbf{Total Drop $\downarrow$}} & \multirow{2}{*}{$\boldsymbol{\pi_{\text{upd}}}$\textbf{(\%)}} \\
 & & Route & Update & (shared) & LLM calls & & \\
\midrule
\multirow{3}{*}{\shortstack[c]{DeepSeek-R1\\7B}}
  & \method{} & \textbf{0} & 106 & 1696 & \textbf{1802} & --- & 2.7 \\
  & Mem0  & 1310 & --- & 1696 & 3006 & 40\% & --- \\
  & $\text{Mem0}\textsuperscript{\textit{g}}$ & 1289 & --- & 1696 & 2985 & 40\% & --- \\
\midrule
\multirow{3}{*}{\shortstack[c]{Qwen2.5\\7B}}
  & \method{} & \textbf{0} & 193 & 1696 & \textbf{1889} & --- & 4.3 \\
  & Mem0  & 1573 & --- & 1696 & 3269 & 42\% & --- \\
  & $\text{Mem0}\textsuperscript{\textit{g}}$ & 1576 & --- & 1696 & 3272 & 42\% & --- \\
\midrule
\multirow{2}{*}{GPT-4o-mini}
  & \method{} & \textbf{0} & 601 & 1696 & \textbf{2297} & --- & 10.3 \\
  & Mem0  & 1521 & --- & 1696 & 3217 & 29\% & --- \\
\bottomrule
\end{tabular}
\vspace{-0.3em}
\caption{
Write-side LLM-call budget on full LoCoMo. \method{} makes \textbf{zero routing calls}, invoking the LLM only to merge the $\pi_{\text{upd}}$ routed to \textsc{Update}, whereas Mem0/Mem0g fuse routing and edit into one call per \texttt{add}. \emph{Total Drop} is \method{}'s reduction of LLM calls vs. the baseline.
}
\label{tab:action-stats}
\vspace{-1em}
\end{table*}

\noindent\textbf{Write-Side Efficiency.}
Table~\ref{tab:action-stats} links these quality gains to a different write-time profile. Within each backbone-matched triad, the dataset, fact-extraction prompt, embedding model, and retrieval stack are fixed: every system issues the same number of fact-extraction LLM calls (one per \texttt{add} call, $1696$ in total), so the only difference is what happens \emph{after} extraction. We therefore separate two layers of cost. ($i$)~At the \emph{decision stage}, Mem0 and $\text{Mem0}\textsuperscript{\textit{g}}$ invoke a routing LLM on every non-empty \texttt{add} call---a single batched call that jointly decides the action and rewrites the memory text for all candidate facts. \method{} instead makes \textbf{zero LLM calls for routing}: the vMF novelty gate resolves \textsc{Add} and \textsc{Noop} in closed form, and the LLM is invoked only to \emph{merge} the small fraction $\pi_{\text{upd}}$ of candidates routed to \textsc{Update}. ($ii$)~Including the shared extraction calls, this yields the \emph{total} write-side LLM budget reported in the last columns.

The two layers tell a deliberately honest story. At the decision stage the reduction is large, around $60$--$90\%$ drop in LLM calls compared to Mem0 on seven of the eight backbones (Table~\ref{tab:action-stats-all}) because \method{} replaces hundreds of routing calls with a handful of merge calls, the empirical update band $\pi_{\text{upd}}$ being narrow ($2.7$--$10.6\%$). Once the shared extraction cost is folded in, the \emph{total} write-side LLM calls still drop by $29$--$42\%$ (mean $32\%$) on those same seven backbones. The single exception is Llama-3.2-1b, where Mem0's weak router emits malformed JSON on $1347$ ($79\%$) of its calls, which artificially lowers its routing-call count rather than reflecting cleaner routing; because SAGE's closed-form gate has no such parse-failure mode, the comparison is not meaningful for this backbone, and we exclude it from the aggregate. 

Read together, Tables~\ref{tab:overall-headtohead}--\ref{tab:action-stats} support the central claim that novelty detection is an effective abstraction for memory evolution. \method{} does not trade quality for efficiency: the same closed-form decision that removes the LLM from clearly novel and clearly redundant candidates also enforces cleaner write-side separation between related-but-distinct facts, which the consistent $F_1$ lead and the multi-hop and open-domain $J$ gains reflect. Efficiency and quality are therefore two faces of a single gating decision rather than competing objectives.
\paragraph{Scaling to a frontier backbone.} The seven small backbones isolate the controller from backbone quality; we now ask whether the same gate holds up on a stronger model by running full LoCoMo on GPT-4o-mini (last block of Table~\ref{tab:overall-headtohead}; Table~\ref{tab:locomo-efficiency}). On quality, \method{} wins multi-hop on $F_1$ and $J$ ($J$ 56.1 {\em vs.} 52.3, +3.7), the category that most directly tests whether the memory system can compose separately stored facts, and edges open-domain $J$ (63.3 {\em vs.} 62.9). Mem0 leads single-hop ($J$ 53.9 {\em vs.} 56.0) and, most clearly, temporal ($J$ 35.4 {\em vs.} 42.7). The overall average $J$ gap is 1.3 points (52.2 {\em vs.} 53.5). The efficiency side is decisive (Table~\ref{tab:locomo-efficiency}): on the same workload \method{} ingests $2.5\times$ faster ($15.7$ {\em vs.}\ $39.3$ min) with $2.6\times$ fewer total write-side tokens ($2.16$M {\em vs.}\ $5.55$M) and $11.1\times$ fewer \emph{generated} tokens ($0.08$M {\em vs.}\ $0.91$M), because the vMF gate replaces Mem0's per-\texttt{add} update-reasoning call with a closed-form vector decision and queries the LLM only on the narrow \textsc{Update} band. Average per-\texttt{add}-call latency is $3.1\times$ lower ($1.76$s {\em vs.}\ $5.38$s) and add-phase API cost falls from \$1.24 to \$0.36 ($3.4\times$ cheaper). These bounded single-hop and temporal recall costs (about $1.3$ average $J$ points)
are thus a deliberate trade for a multiplicative reduction in write-side compute that only compounds as the corpus grows.

\begin{table}[t]
\vspace{-1em}
\centering
\small
\begin{tabular}{l r r}
\toprule
Metric & Mem0 & SAGE \\
\midrule
Write/add LLM calls & 3,217 & 2,297 \\
Add prompt tokens & 4,637,649 & 2,072,575 \\
Add completion tokens & 913,982 & 82,632 \\
Add total tokens & 5,551,631 & 2,155,207 \\
Avg.\ latency / add call (s) & 5.38 & 1.76 \\
Add wall-clock (min) & 39.3 & 15.7 \\
Search wall-clock (min) & 66.7 & 65.5 \\
Total wall-clock (min) & 106.3 & 81.5 \\
Add API cost (USD, 4o-mini) & \$1.24 & \$0.36 \\
\bottomrule
\end{tabular}
\caption{Efficiency on full LoCoMo (w/ GPT-4o-mini). Add-phase token counts and per-call latency are measured at the API boundary. }
\vspace{-1em}
\label{tab:locomo-efficiency}
\end{table}

\begin{table}[t]
\centering
\small
\setlength{\tabcolsep}{3.5pt}
\begin{tabular}{lccrr}
\toprule
\textbf{Model} & \textbf{Skip} & \textbf{Calls} & \multicolumn{1}{c}{$\boldsymbol{\Delta} F_1$} & \multicolumn{1}{c}{$\boldsymbol{\Delta} J$} \\
 & \textbf{rate} & \textbf{saved} & \multicolumn{1}{c}{(\%)} & \multicolumn{1}{c}{(\%)} \\
\midrule
Qwen2.5-3B  & 16.8\% & $1{,}946$ & $+0.05$           & $-0.50$ \\
Qwen2.5-7B  & 17.4\% & $2{,}016$ & $-1.27$   & $-0.20$ \\
Llama-3.1-8B             & 15.8\% & $1{,}824$       & $-1.44$           & $-0.15$ \\
Llama-3.2-3B             & 16.4\% & $1{,}896$       & $-0.01$           & $+0.65$ \\
\midrule
GPT-4o-mini              & 17.9\% & $2{,}066$       & $-0.53$           & $-2.01$ \\
\bottomrule
\end{tabular}
\caption{Fixed-threshold NOOP decision ($\tau_{\text{noop}}=0.572$): A-Mem+\method{} compared to A-Mem baseline on full LoCoMo, in percentage points ($F_1$, $J$). ``Calls saved'' = skipped write/evolution LLM calls.}
\vspace{-1.5em}
\label{tab:leakfree-crossfamily}
\end{table}

\subsection{Threshold Sensitivity Ablation}
\label{sec:adaptive-threshold}

To analyze the adaptive threshold sensitivity, we compare \method{} with adaptive threshold $\tau_t$ against \method{} with $\tau_t$ set to fixed thresholds, say, \(\tau_{\text{fixed}} \in \{0.10,0.15,0.20,0.25,0.30\}\) using a 20\% subsample of LoCoMo, and Llama-3.1-8B as the LLM judge. 
The results in Appendix Table~\ref{tab:vmf-threshold-overall} show that adaptive \method{} is the more robust default operating point. On Qwen2.5-1.5B, it gives the best overall $B_1$ (\(9.80\)) and $F_1$ (\(11.69\)); the only fixed threshold that slightly exceeds its $J$ score is \(\tau_{\text{fixed}}=0.30\), and only by \(0.07\), while $B_1$ and $F_1$ both drop by about \(2\) points. On Qwen2.5-3B, the best fixed point is \(\tau_{\text{fixed}}=0.10\), which improves $B_1$ from \(25.83\) to \(26.69\), $F_1$ from \(31.15\) to \(32.35\), and $J$ from \(85.32\) to \(86.82\).

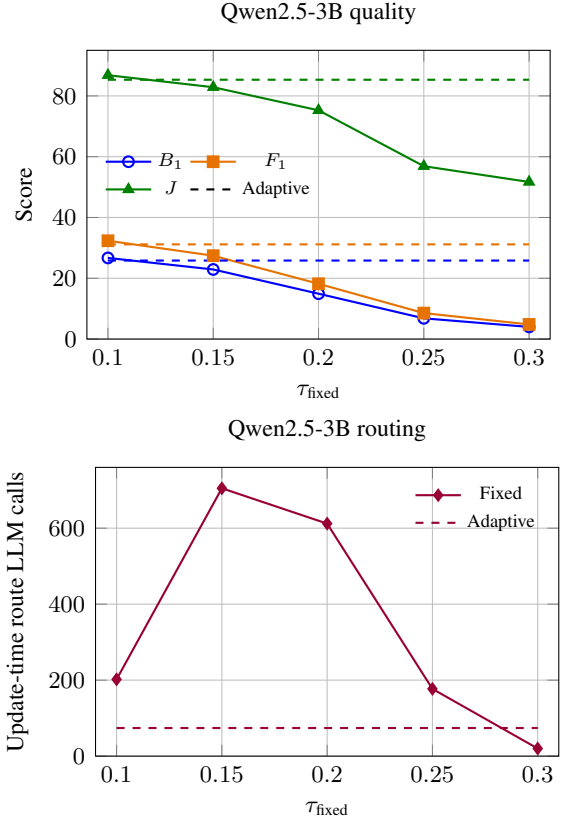
\begin{figure}[t]
\centering
\begin{minipage}[t]{\linewidth}
\centering
\begin{tikzpicture}
\begin{axis}[
width=\linewidth,
height=5.4cm,
xlabel={\(\tau_{\text{fixed}}\)},
ylabel={Score},
xmin=0.09, xmax=0.31,
ymin=0.0, ymax=95,
xtick={0.10,0.15,0.20,0.25,0.30},
grid=major,
legend style={at={(0.02,0.45)}, anchor=south west, font=\scriptsize, draw=none, fill=none},
legend columns=2,
tick label style={font=\small},
label style={font=\small},
title style={font=\small},
title={Qwen2.5-3B quality},
]
\addplot[blue, mark=o, thick] coordinates {(0.10,26.69) (0.15,22.90) (0.20,14.92) (0.25,6.82) (0.30,3.99)};
\addlegendentry{\(B_1\)}
\addplot[orange!90!black, mark=square*, thick] coordinates {(0.10,32.35) (0.15,27.43) (0.20,18.19) (0.25,8.54) (0.30,4.81)};
\addlegendentry{\(F_1\)}
\addplot[green!50!black, mark=triangle*, thick] coordinates {(0.10,86.82) (0.15,82.86) (0.20,75.26) (0.25,56.88) (0.30,51.69)};
\addlegendentry{\(J\)}
\addplot[blue, dashed, thick, forget plot] coordinates {(0.10,25.83) (0.30,25.83)};
\addplot[orange!90!black, dashed, thick, forget plot] coordinates {(0.10,31.15) (0.30,31.15)};
\addplot[green!50!black, dashed, thick, forget plot] coordinates {(0.10,85.32) (0.30,85.32)};
\addlegendimage{black, dashed, thick}
\addlegendentry{Adaptive}
\end{axis}
\end{tikzpicture}
\end{minipage}

\begin{minipage}[t]{\linewidth}
\centering
\begin{tikzpicture}
\begin{axis}[
width=\linewidth,
height=5.4cm,
xlabel={\(\tau_{\text{fixed}}\)},
ylabel={Update-time route LLM calls},
xmin=0.09, xmax=0.31,
ymin=0, ymax=760,
xtick={0.10,0.15,0.20,0.25,0.30},
grid=major,
tick label style={font=\small},
label style={font=\small},
title style={font=\small},
title={Qwen2.5-3B routing},
legend style={at={(0.67,0.74)}, anchor=south west, font=\scriptsize, draw=none, fill=none},
legend columns=1,
]
\addplot[purple!80!black, mark=diamond*, thick] coordinates {(0.10,202) (0.15,705) (0.20,612) (0.25,177) (0.30,20)};
\addlegendentry{Fixed}
\addplot[purple!80!black, dashed, thick, forget plot] coordinates {(0.10,74) (0.30,74)};
\addlegendimage{purple!80!black, dashed, thick}
\addlegendentry{Adaptive}
\end{axis}
\end{tikzpicture}
\end{minipage}
\caption{Adaptive threshold sensitivity on Qwen2.5-3B. Left: quality under fixed thresholds. Right: update-time route LLM calls, where only update-routed candidates invoke the LLM call. Solid lines indicate \method{} with varying fixed-threshold and dashed lines indicate \method{} with adaptive threshold. 
}
\vspace{-1em}
\label{fig:paper-threshold-quality}
\end{figure}

Figure~\ref{fig:paper-threshold-quality} shows the
same trade-off as Appendix Table~\ref{tab:vmf-threshold-overall}: the best fixed quality point is $\tau_{\text{fixed}} = 0.10$, but adaptive SAGE stays close while
using far fewer update-time calls: useful operating points are concentrated in the low-threshold region, and quality degrades sharply once \(\tau_{\text{fixed}} \ge 0.15\). The right panel also makes the efficiency trade-off explicit: the best fixed quality point uses nearly \(3\times\) as many update-time route calls as adaptive \method{} (202 {\em vs.}\ 74). The broader threshold-sensitivity pattern across both backbones appears in Appendix Figure~\ref{fig:paper-threshold-routing}. Overall, the adaptive controller already captures most of the attainable quality without backbone-specific retuning, which is the practical significance of \method{} as a write-time control policy.

\subsection{Isolating \textsc{Noop} Decision's Effects}
\label{sec:leakfree-exp}
To isolate the \textsc{Noop} decision, we hold the underlying A-Mem memory system fixed and change only whether the fixed-threshold gate of Section~\ref{sec:leakfree} is switched \emph{on} or \emph{off}. Therefore, any difference between the two methods A-Mem and A-Mem+\method{} is attributable to \method{} alone. Table~\ref{tab:leakfree-crossfamily} reports the result across five models on full LoCoMo 
with threshold $\tau_{\text{noop}}=0.572$ calculated and fixed in advance (details of how to set $\tau_{\text{noop}}$ are in Appendix~\ref{app:leakfree-calib}). Read across Table~\ref{tab:leakfree-crossfamily}, the gate behaves as designed. The \emph{skip-rate} column lands in $15.8$--$17.9\%$ for every model. Each run therefore avoids $1{,}824$--$2{,}066$ write/evolution LLM calls (\emph{calls-saved} column). The \emph{$\Delta J$} score gain column shows this efficiency is essentially free on the four open-weight models: $J$ shifts by at most $0.65\%$ in either direction ($\le 1$ point), (per-category breakdown in Appendix~\ref{app:add_res}). At a comparable $17.9\%$ skip rate, \method{} costs $2.01\%$ in $J$ for GPT-4o-mini model.

\section{Conclusion}
This paper argues that novelty detection is the missing abstraction for write-side memory control in agentic LLMs. Prior systems have shown that memory evolution matters, but they typically rely on controller LLMs to decide whether a new fact should trigger \textsc{Add}, \textsc{Update}, or \textsc{Noop} behavior. We instead propose \method{}, a von Mises--Fisher novelty gate, which yields a simple operational principle: add clearly novel memories, ignore clearly redundant ones, and reserve local merge reasoning for the uncertainty band in between.

\section*{Limitations}
Our evaluation is conducted entirely on the LoCoMo benchmark in English, covering one interaction modality (multi-session dialogue). We have not tested \method{} on harder benchmarks such as LongMemEval, on task-oriented or tool-use agent settings, or on multilingual corpora, so the generality of the quality--efficiency trade-off remains open. The gate routes candidates to \textsc{Add}, \textsc{Update}, or \textsc{Noop} but does not issue \textsc{Delete} decisions, nor does the current system include a memory compaction mechanism; designing principled deletion and compaction strategies that integrate with the vMF novelty score is left to future work. Finally, because the vMF score operates on $\ell_2$-normalized sentence embeddings, it inherits the embedding model's limitations: semantically distinct facts that receive similar vectors may be incorrectly dropped, while paraphrases with dissimilar vectors may bypass the redundancy filter. 

\bibliography{references}

\clearpage

\appendix

\section{Dataset: LoCoMo}
\label{app:locomo-dataset}
All experiments use the LoCoMo benchmark~\citep{maharana2024evaluating}, which targets long-horizon conversational memory. The corpus consists of 10 multi-session dialogues in which two speakers share and revisit personal experiences over an extended interaction history. Each dialogue spans roughly 600 turns ($\approx$26k tokens) and is paired with around 200 post-hoc comprehension questions whose ground-truth answers require the system to recall facts from the conversation.
We adopt the four question categories relevant to memory-write quality: \emph{single-hop} questions that probe a single stored fact, \emph{multi-hop} questions that require composing information across turns or sessions, \emph{temporal} questions that test sensitivity to the ordering or timing of events, and \emph{open-domain} questions that additionally draw on commonsense knowledge.
The original benchmark also defines an adversarial category, but ground-truth answers are not provided for these questions and the expected system behavior is to recognize them as unanswerable~\citep{mem0,xu2025mem}. Because this tests abstention rather than memory-write fidelity, we exclude it from our evaluation.

\section{Baseline Descriptions}
\label{app:baselines}

\paragraph{Mem0.}
Mem0~\citep{mem0} is a memory layer for LLM agents that extracts salient facts from conversation turns and maintains them in a dense vector store. For each candidate fact, an LLM-based routing controller inspects the top-$k$ most similar existing memories and classifies the appropriate operation as one of \textsc{Add}, \textsc{Update}, \textsc{Delete}, or \textsc{Noop}. This design enables compact natural-language memory representations---averaging roughly 7k tokens per conversation on LoCoMo---but requires one routing LLM call per batch of extracted candidates at every write step, making the write-time cost proportional to the total number of ingested turns regardless of their novelty. Retrieval is performed via cosine similarity over the dense embedding index.

\paragraph{Mem0\textsuperscript{\textit{g}}.}
Mem0\textsuperscript{\textit{g}}~\citep{mem0} extends Mem0 with a graph-based memory layer stored in Neo4j. An LLM-driven extraction pipeline converts conversation messages into typed entity nodes and directed relation triplets of the form $(v_s, r, v_d)$. When new triplets arrive, the system computes entity embeddings, searches for semantically similar existing nodes above a similarity threshold, and applies a conflict-detection and update-resolution mechanism via additional LLM calls to maintain graph consistency. At query time, Mem0\textsuperscript{\textit{g}} employs a dual retrieval strategy: an entity-centric method that traverses the graph neighborhood of query-matched nodes, and a semantic-triplet method that matches the full query embedding against all stored triplet encodings. The graph layer roughly doubles the token footprint relative to Mem0 (approximately 14k tokens per conversation) but provides gains on temporal and open-domain questions where relational structure is beneficial.

\paragraph{A-Mem.}
A-Mem~\citep{xu2025mem} is an agentic memory system inspired by the Zettelkasten method that organises memories as interconnected atomic notes. Each note stores the original content alongside LLM-generated keywords, tags, and a contextual description, all concatenated into a single embedding for similarity search. Upon insertion, the system retrieves the top-$k$ nearest existing notes and prompts an LLM to determine whether semantic links should be established; linked notes are grouped into overlapping ``boxes'' that are co-retrieved at query time. A-Mem further implements a \emph{memory evolution} step: when a new note is integrated, the LLM may rewrite the contextual descriptions and attributes of its linked neighbours, enabling the memory network to refine its organisation over time. While A-Mem reduces retrieval-time token budgets to roughly 1.2--2.5k tokens, it still issues multiple LLM calls per insertion (note construction, link generation, and evolution), placing the bulk of its computational cost on the write side.

\section{Additional Hyperparameter and Experimental Details}
\label{app:hyperparameter-details}

The adaptive routing rule in \method{} (Section~\ref{sec:adaptive-routing}) has three core parameters that govern the density-dependent threshold $\tau_t^\star = \tau_{\min} + \tau_0 \, e^{-\lambda \rho_t}$: the base scaling parameter $\tau_0$, the minimum threshold floor $\tau_{\min}$, and the density decay coefficient $\lambda$. We selected these via a grid search on Qwen2.5-3B using a 20\% subsample of LoCoMo, sweeping over
$\tau_0 \in \{0.15,\, 0.25,\, 0.35\}$, $\tau_{\min} \in \{0.01,\, 0.025,\, 0.05\}$,
$\lambda \in \{1.0,\, 2.0,\, 4.0\}$.
The configuration $\tau_0 = 0.25$, $\tau_{\min} = 0.025$, $\lambda = 2.0$ was selected and held fixed across all eight backbones reported in the paper. No per-backbone retuning was performed.

The remaining parameters serve different roles and were set without search.
The EMA momentum $\alpha = 0.9$ smooths the threshold across consecutive write steps so that a single conversational turn that adds several memories does not cause an abrupt shift in the decision boundary; the specific value reflects a standard smoothing rate and was not tuned, but it is consistent with standard EMA based updates.
The uncertainty-band half-width $\delta = 0.025$ controls how many candidates are routed to the \textsc{Update} path and thereby how many expensive LLM merge calls are issued at write time. Increasing $\delta$ widens the band and sends more borderline candidates to the LLM for deliberation; decreasing it narrows the band and favors the cheaper \textsc{Add}/\textsc{Noop} decisions. In practice, $\delta$ can therefore be treated as an operational knob that trades update quality against write-side compute.
The PCA projection dimension is set to $d' = 16$ only affects the density proxy (Appendix~\ref{app:proxyrho}).

All the experiments are run using an NVIDIA H200 GPU, and one single run completes in around 2 hours for larger models. 
The code used Python 3.9.25, PyTorch 2.4.0, and NLTK 3.9.2.

\section{Proxy for Memory Scope Density}\label{app:proxyrho}
At write step $t$, let $\mathcal{M}^{(t)} = \{\mathbf{m}^{(t)}_{1}, \ldots, \mathbf{m}^{(t)}_{N_t}\}$ denote the current memory scope. 
To estimate geometric spread, we project the memory vectors onto their first \(d'\) principal components, obtaining \(u_i^{(t)} \in \mathbb{R}^{d'}\)~\cite{pearson1901liii}. This lets us measure spread using the main directions of variation in the memory vectors, while avoiding noisy range estimates in dimensions where the vectors change very little.
We then define the scope volume as the product of the coordinate-wise ranges in this projected space,
\begin{equation}
V_t=\exp\left(\sum_{j=1}^{d'} \log (\max_i u_{i,j}^{(t)}-\min_i u_{i,j}^{(t)})\right),
\end{equation}
where \(u_{i,j}^{(t)}\) is the \(j\)-th coordinate of the \(i\)-th projected memory at step \(t\). Intuitively, \(V_t\) is large when the current memories are spread out across the informative directions and small when they are tightly packed. Thus, we form the following approximation for the density proxy:
\begin{equation}
\rho_t=\frac{N_t}{V_t}.
\end{equation}
When $\rho_t$ is large, the memory store contains many items within a small effective volume of the projected subspace.  As a result, neighborhood support becomes easier to accumulate: incoming candidates are more likely to lie close to already populated regions, which systematically depresses their novelty scores.  If the threshold were kept fixed, the controller would become overly conservative in dense stores and would reject too many genuinely useful writes; accordingly, the gate should lower its threshold and become more permissive as density increases.
\section{Bound on the vMF Aggregation Score}
\label{app:vmf-score-bound}

We restate and prove the bound used in Section~\ref{sec:methodsage}.

\paragraph{Proposition.}
Let \(\mathcal{M} = \{\mathbf{m}_1,\dots,\mathbf{m}_N\} \subset \mathbb{S}^{d-1}\) be a nonempty memory scope with \(N \ge 1\), where
\[
\mathbb{S}^{d-1} = \{\mathbf{z} \in \mathbb{R}^d : \|\mathbf{z}\|_2 = 1\}
\]
is the unit hypersphere in \(\mathbb{R}^d\). Let \(\mathbf{c} \in \mathbb{S}^{d-1}\) be a candidate embedding, and let \(\kappa > 0\) denote the concentration parameter. We define
\begin{align*}
K_\kappa(\mathbf{c},\mathbf{m}_i)=\exp(\kappa\,\mathbf{m}_i^\top \mathbf{c}),\\
\hat{S}(\mathbf{c}\mid\mathcal{M})=\frac{1}{N}\sum_{i=1}^{N}K_\kappa(\mathbf{c},\mathbf{m}_i),
\end{align*}
and
\begin{equation}
s_{\mathrm{vMF}}(\mathbf{c}\mid\mathcal{M})
=
\frac{1}{\kappa}
\log \hat{S}(\mathbf{c}\mid\mathcal{M}).
\end{equation}
Then
\begin{equation}
-1 \le s_{\mathrm{vMF}}(\mathbf{c}\mid\mathcal{M}) \le 1.
\end{equation}

\paragraph{Proof.}
Since \(\mathbf{c},\mathbf{m}_i \in \mathbb{S}^{d-1}\), we have
\begin{align*}
&\|\mathbf{c}\|_2 = 1
\qquad\text{and}\qquad
\|\mathbf{m}_i\|_2 = 1 \\
&\text{for all } i=1,\dots,N.
\end{align*}
Therefore, by the Cauchy--Schwarz inequality,
\[
|\mathbf{m}_i^\top \mathbf{c}|
\le
\|\mathbf{m}_i\|_2 \,\|\mathbf{c}\|_2
=
1,
\]
which implies
\[
-1 \le \mathbf{m}_i^\top \mathbf{c} \le 1
\quad \text{for all } i=1,\dots,N.
\]

Because \(\kappa > 0\), multiplying by \(\kappa\) preserves the inequality:
\[
-\kappa \le \kappa\,\mathbf{m}_i^\top \mathbf{c} \le \kappa
\quad \text{for all } i=1,\dots,N.
\]
 
By the definition of \(K_\kappa\) and the monotonicity of the exponential function,
\[
e^{-\kappa}
\le
K_\kappa(\mathbf{c},\mathbf{m}_i)
\le
e^{\kappa}
\quad \text{for all } i=1,\dots,N.
\]
 
Since \(\hat{S}(\mathbf{c}\mid\mathcal{M})\) is the arithmetic mean of these \(N\) terms, averaging over \(i=1,\dots,N\) gives
\[
e^{-\kappa}
\le
\hat{S}(\mathbf{c}\mid\mathcal{M})
\le
e^{\kappa}.
\]
 
Applying the logarithm, which is also monotone increasing on \((0,\infty)\), yields
\[
-\kappa
\le
\log \hat{S}(\mathbf{c}\mid\mathcal{M})
\le
\kappa.
\]
 
Finally, dividing by \(\kappa > 0\) gives
\[
-1
\le
\frac{1}{\kappa}
\log \hat{S}(\mathbf{c}\mid\mathcal{M})
\le
1.
\]
Hence,
\[
-1 \le s_{\mathrm{vMF}}(\mathbf{c}\mid\mathcal{M}) \le 1.
\]
\qed

\section{Temporal Dynamics of the Adaptive Threshold}
\label{app:illustrateadapt}

\begin{figure}[t]
\centering
\includegraphics[width=0.45\textwidth]{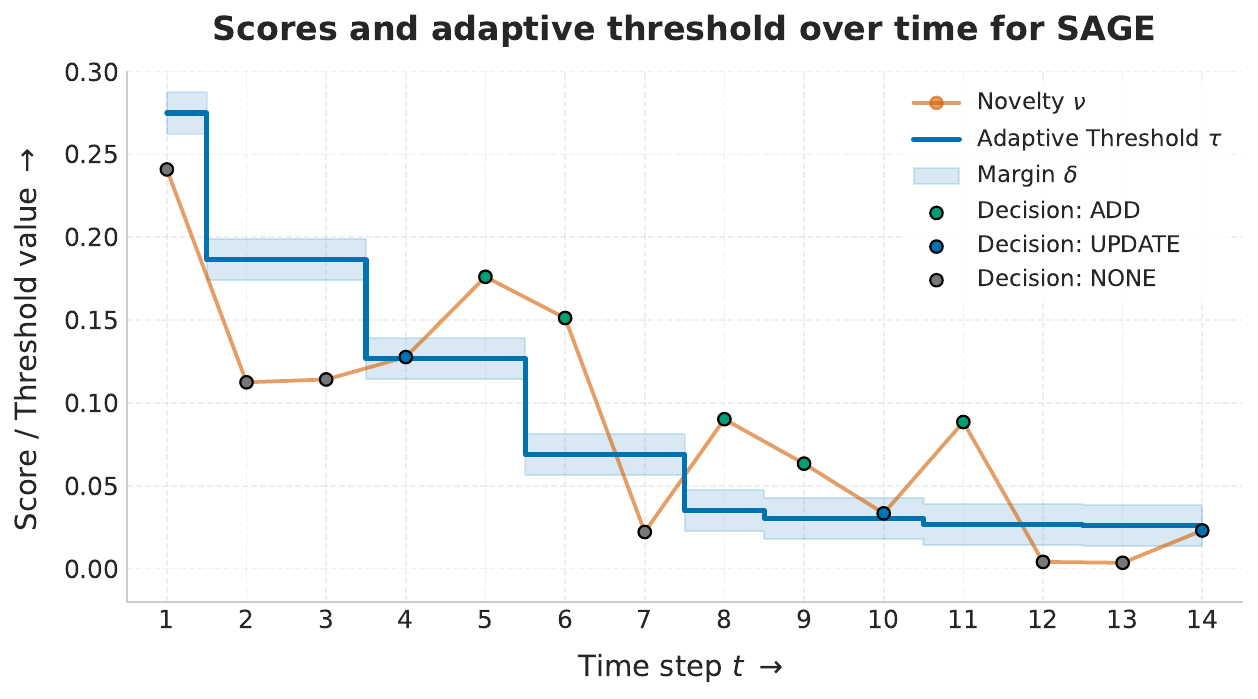}
\caption{Illustration of decaying adaptive threshold over time that influences routing decisions for \method{}.
}
\label{fig:illustrateadapt}
\end{figure}

Figure \ref{fig:illustrateadapt} provides a step-by-step visual trace of the \method{} routing mechanism in action across a sequence of candidate facts. As the memory scope expands and the projection subspace becomes more densely populated, the baseline novelty scores of incoming candidates naturally trend downward because new facts are more likely to fall near established memories. To prevent the system from becoming overly conservative, the adaptive threshold $\tau_t$ (the solid blue line) decays over time in response to the increasing density proxy $\rho_t$.

The figure illustrates how the uncertainty margin $\delta$ (the shaded blue band above the threshold) cleanly separates the three routing actions defined in Section \ref{sec:methodology}:

\begin{itemize}[leftmargin=*]
    \item \textbf{\textsc{Add}:} Candidates landing strictly above the shaded band ($\nu(\mathbf{c}) \ge \tau_t + \delta$).
    \item \textbf{\textsc{Update}:} Candidates landing inside the shaded band ($\tau_t \le \nu(\mathbf{c}) < \tau_t + \delta$).
    \item \textbf{\textsc{Noop}:} Candidates scoring strictly below the threshold ($\nu(\mathbf{c}) < \tau_t$).
\end{itemize}

By continuously shifting downward as memory density increases, this dynamic adjustment ensures that the decision boundary remains correctly calibrated to the current state of the memory store, preserving high recall without sacrificing write-time efficiency.

\section{Leakage-Controlled Calibration of \texorpdfstring{$\tau_{\text{noop}}$}{tau\_noop}}
\label{app:leakfree-calib}
This appendix details how the operating point $\tau_{\text{noop}}$ of the \textsc{Noop} gate (Section~\ref{sec:leakfree}) is fixed for a new deployment without consulting the target benchmark.

\noindent\textbf{The calibration trap.} The tempting recipe is to read the threshold off the benchmark: compute the $s_{\text{vMF}}$ distribution on LoCoMo and place $\tau_{\text{noop}}$ at its $80$th percentile, so the gate skips the most-covered $20\%$ of writes (on LoCoMo this percentile is $0.572$). Although unsupervised---it never touches the labels---this is still \emph{test-set calibration}: a hyperparameter of the evaluated method is read off the evaluation distribution, which a real deployment does not have in advance.

\noindent\textbf{Leakage-controlled calibration.} We instead fix $\tau_{\text{noop}}$ offline, on synthetic self-generated text that never sees LoCoMo, constructed so that its $s_{\text{vMF}}$ distribution \emph{matches} that of real conversational memory. Once a synthetic corpus reproduces the real $80$th-percentile score, the rule ``skip the top $20\%$ most-covered'' yields the same threshold value---now a property of our recipe rather than of the benchmark. The effective lever is \emph{topical coherence}, not surface naturalness: rigid templated text over-concentrates ($p_{80}=0.892$, everything looks redundant), whereas a broad LLM-generated life story is too topically diffuse ($p_{80}=0.443$, everything looks novel). A narrow-domain, single-persona diary lands on the real spread, with generation temperature acting as a clean monotonic knob on synthetic redundancy (Table~\ref{tab:leakfree-temp}). At temperature $0.7$ the synthetic $p_{80}$ matches LoCoMo's $0.572$, which we adopt as $\tau_{\text{noop}}=0.572$. The corpus, the chosen quantile, and the threshold value are thus all derived from synthetic data alone, making the threshold transferable rather than tuned to the benchmark.

\begin{table}[t]
\centering
\small
\begin{tabular}{ccc}
\toprule
Gen.\ temperature & Synthetic $p_{80}$ & LoCoMo skip @ $p_{80}$ \\
\midrule
0.5 & 0.587 & 15.2\% \\
\textbf{0.7} & \textbf{0.572} & \textbf{20.0\%} \\
0.9 & 0.543 & 31.5\% \\
\bottomrule
\end{tabular}
\caption{Leakage-controlled threshold calibration. A narrow-domain, single-persona synthetic diary corpus reproduces LoCoMo's $80$th-percentile vMF score at generation temperature $0.7$, giving $\tau_{\text{noop}}=0.572$ with no benchmark access. Temperature is a clean monotonic knob on synthetic redundancy.}
\label{tab:leakfree-temp}
\end{table}

\begin{table*}[ht]
\centering
\small
\setlength{\tabcolsep}{6pt}
\renewcommand{\arraystretch}{0.9}
\begin{tabular}{llrrr}
\toprule
\textbf{Model} & \textbf{Variant} & \textbf{BLEU-1} & \textbf{Token-$F_1$} & \textbf{Judge} \\
\midrule
\multirow{6}{*}{\shortstack[l]{qwen2.5-1.5b}} & adaptive & \textbf{9.80} & \textbf{11.69} & 71.62 \\
 & $\tau_{\text{fixed}}=0.10$ & 8.80 & 10.94 & 69.81 \\
 & $\tau_{\text{fixed}}=0.15$ & 8.76 & 10.83 & 70.58 \\
 & $\tau_{\text{fixed}}=0.20$ & 8.64 & 10.51 & 70.91 \\
 & $\tau_{\text{fixed}}=0.25$ & 8.01 & 9.81 & 69.29 \\
 & $\tau_{\text{fixed}}=0.30$ & 7.81 & 9.77 & \textbf{71.69} \\
\midrule
\multirow{6}{*}{\shortstack[l]{qwen2.5-3b}} & adaptive & 25.83 & 31.15 & 85.32 \\
 & $\tau_{\text{fixed}}=0.10$ & \textbf{26.69} & \textbf{32.35} & \textbf{86.82} \\
 & $\tau_{\text{fixed}}=0.15$ & 22.90 & 27.43 & 82.86 \\
 & $\tau_{\text{fixed}}=0.20$ & 14.92 & 18.19 & 75.26 \\
 & $\tau_{\text{fixed}}=0.25$ & 6.82 & 8.54 & 56.88 \\
 & $\tau_{\text{fixed}}=0.30$ & 3.99 & 4.81 & 51.69 \\
\bottomrule
\end{tabular}%
\caption{Overall adaptive-vs-fixed threshold ablation for \method{}. Each row uses the paired full-split run scored with \texttt{llama3.1-8b} as the LLM judge. Bold marks the best quality value within the same backbone.}
\label{tab:vmf-threshold-overall}
\end{table*}

\section{Additional Results}
\label{app:add_res}
\subsection{Threshold Ablation Details}
\label{app:threshold-ablation}

This appendix provides the full adaptive-vs-fixed threshold sweep referenced in Section~\ref{sec:adaptive-threshold}.

\paragraph{Table~\ref{tab:vmf-threshold-overall}: overall quality under fixed vs.\ adaptive thresholds.}
Each row reports the overall BLEU-1, token-$F_1$, and LLM-as-a-Judge score when \method{} is run with either its adaptive threshold $\tau_t$ or a fixed value $\tau_{\text{fixed}} \in \{0.10, 0.15, 0.20, 0.25, 0.30\}$, scored on a 20\% subsample of LoCoMo with Llama-3.1-8B as the judge.

Two patterns emerge:
{\em (i)}~On Qwen2.5-1.5B, the adaptive threshold yields the best $B_1$ ($9.80$) and $F_1$ ($11.69$) across all settings; the only fixed threshold that slightly exceeds its Judge score is $\tau_{\text{fixed}}=0.30$ (by $0.07$ points), but at a cost of roughly $2$ points on both $B_1$ and $F_1$.
{\em (ii)}~On Qwen2.5-3B, the best fixed setting ($\tau_{\text{fixed}}=0.10$) slightly outperforms the adaptive threshold on all three metrics ($B_1$: $26.69$ vs.\ $25.83$; $F_1$: $32.35$ vs.\ $31.15$; $J$: $86.82$ vs.\ $85.32$), but quality degrades sharply for $\tau_{\text{fixed}} \ge 0.15$ and collapses by $\tau_{\text{fixed}} = 0.30$ ($F_1$ drops to $4.81$).
Because the adaptive threshold performs well across both backbones without requiring per-backbone tuning, it is the more robust default.

\begin{figure}[ht]
\centering
\begin{minipage}[t]{\linewidth}
\centering
\begin{tikzpicture}
\begin{axis}[
width=\linewidth,
height=4.6cm,
xlabel={\(\tau_{\text{fixed}}\)},
ylabel={Score},
xmin=0.09, xmax=0.31,
ymin=0.0, ymax=95,
xtick={0.10,0.15,0.20,0.25,0.30},
grid=major,
legend style={at={(0.97,0.5)}, anchor=east, font=\scriptsize, draw=none, fill=none},
legend columns=1,
tick label style={font=\small},
label style={font=\small},
title style={font=\small},
title={Qwen2.5-1.5B},
]
\addplot[blue, mark=o, thick] coordinates {(0.10,8.80) (0.15,8.76) (0.20,8.64) (0.25,8.01) (0.30,7.81)};
\addlegendentry{BLEU-1}
\addplot[orange!90!black, mark=square*, thick] coordinates {(0.10,10.94) (0.15,10.83) (0.20,10.51) (0.25,9.81) (0.30,9.77)};
\addlegendentry{Token-\(F_1\)}
\addplot[green!50!black, mark=triangle*, thick] coordinates {(0.10,69.81) (0.15,70.58) (0.20,70.91) (0.25,69.29) (0.30,71.69)};
\addlegendentry{Judge}
\addplot[blue, dashed, thick, forget plot] coordinates {(0.10,9.80) (0.30,9.80)};
\addplot[orange!90!black, dashed, thick, forget plot] coordinates {(0.10,11.69) (0.30,11.69)};
\addplot[green!50!black, dashed, thick, forget plot] coordinates {(0.10,71.62) (0.30,71.62)};
\addlegendimage{black, dashed, thick}
\addlegendentry{Adaptive}
\end{axis}
\end{tikzpicture}
\end{minipage}

\vspace{1ex}

\begin{minipage}[t]{\linewidth}
\centering
\begin{tikzpicture}
\begin{axis}[
width=\linewidth,
height=4.6cm,
xlabel={\(\tau_{\text{fixed}}\)},
ylabel={Score},
xmin=0.09, xmax=0.31,
ymin=0.0, ymax=95,
xtick={0.10,0.15,0.20,0.25,0.30},
grid=major,
tick label style={font=\small},
label style={font=\small},
title style={font=\small},
title={Qwen2.5-3B},
]
\addplot[blue, mark=o, thick] coordinates {(0.10,26.69) (0.15,22.90) (0.20,14.92) (0.25,6.82) (0.30,3.99)};
\addplot[orange!90!black, mark=square*, thick] coordinates {(0.10,32.35) (0.15,27.43) (0.20,18.19) (0.25,8.54) (0.30,4.81)};
\addplot[green!50!black, mark=triangle*, thick] coordinates {(0.10,86.82) (0.15,82.86) (0.20,75.26) (0.25,56.88) (0.30,51.69)};
\addplot[blue, dashed, thick, forget plot] coordinates {(0.10,25.83) (0.30,25.83)};
\addplot[orange!90!black, dashed, thick, forget plot] coordinates {(0.10,31.15) (0.30,31.15)};
\addplot[green!50!black, dashed, thick, forget plot] coordinates {(0.10,85.32) (0.30,85.32)};
\end{axis}
\end{tikzpicture}
\end{minipage}
\caption{Threshold sensitivity to the fixed threshold across both Qwen backbones. Colors denote BLEU-1, token-$F_1$, and Judge; solid lines indicate \method{} with varying fixed-thresholds and dashed lines indicate \method{} with adaptive threshold.}
\label{fig:paper-threshold-routing}
\end{figure}
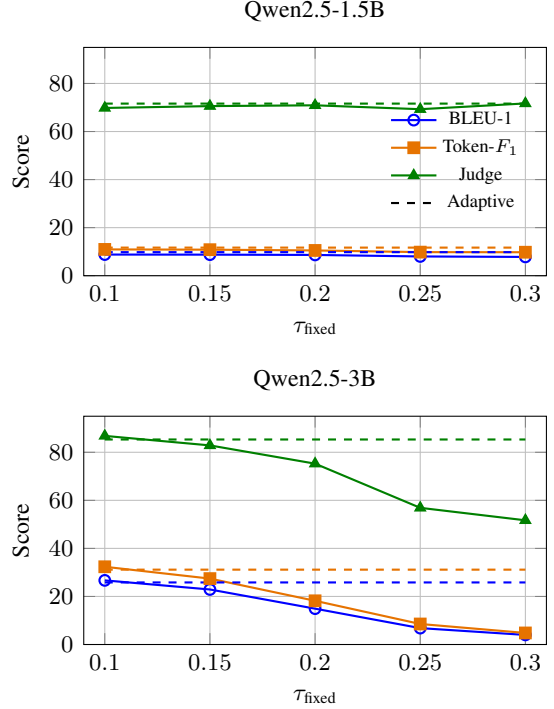

\paragraph{Figure~\ref{fig:paper-threshold-routing}: threshold-sensitivity curves across both backbones.}
Figure~\ref{fig:paper-threshold-routing} visualizes the same data as Table~\ref{tab:vmf-threshold-overall} in plot form. Solid lines trace the three quality metrics as a function of $\tau_{\text{fixed}}$; dashed horizontal lines mark the corresponding adaptive-\method{} baselines.
On Qwen2.5-1.5B (top panel), all three solid curves remain relatively flat, never clearly exceeding the adaptive baselines, confirming that no single fixed threshold consistently dominates the adaptive gate on this backbone.
On Qwen2.5-3B (bottom panel), the curves are steeply right-descending: $\tau_{\text{fixed}}=0.10$ is the only competitive operating point, and every higher threshold incurs a severe quality penalty. This asymmetry highlights the fragility of fixed thresholds, as the optimal $\tau_{\text{fixed}}$ shifts across backbones, whereas the adaptive threshold automatically tracks memory-store geometry and remains robust across configurations.

\begin{table*}[t]
\centering
\small
\setlength{\tabcolsep}{6pt}
\renewcommand{\arraystretch}{0.9}
\begin{tabular}{cc ccccccc}
\toprule
\multirow{2}{*}{\textbf{Model}} & \multirow{2}{*}{\textbf{System}} & \multicolumn{2}{c}{\textbf{Decision-Stage LLM}} & \textbf{Extraction} & \textbf{Total write} & \textbf{Total} & \multirow{2}{*}{$\boldsymbol{\pi_{\text{upd}}}$\textbf{(\%)}} \\
 & & Route & Update & (shared) & LLM calls & $\downarrow$ vs.\ row & \\
\midrule
\multirow{3}{*}{DeepSeek-R1-1.5b}
  & \method{} & \textbf{0} & 115 & 1696 & \textbf{1811} & --- & 3.2 \\
  & Mem0  & 872 & --- & 1696 & 2568 & 29\% & --- \\
  & $\text{Mem0}\textsuperscript{\textit{g}}$ & 895 & --- & 1696 & 2591 & 30\% & --- \\
\midrule
\multirow{3}{*}{\shortstack[c]{DeepSeek-R1\\7B}}
  & \method{} & \textbf{0} & 106 & 1696 & \textbf{1802} & --- & 2.7 \\
  & Mem0  & 1310 & --- & 1696 & 3006 & 40\% & --- \\
  & $\text{Mem0}\textsuperscript{\textit{g}}$ & 1289 & --- & 1696 & 2985 & 40\% & --- \\
\midrule
\multirow{3}{*}{\shortstack[c]{Llama-3.2$^\dagger$\\1B}}
  & \method{} & \textbf{0} & 397 & 1696 & 2093 & --- & 4.8 \\
  & Mem0  & 349 & --- & 1696 & 2045 & $-2$\% & --- \\
  & $\text{Mem0}\textsuperscript{\textit{g}}$ & 402 & --- & 1696 & 2098 & 0\% & --- \\
\midrule
\multirow{3}{*}{\shortstack[c]{Llama-3.2\\3B}}
  & \method{} & \textbf{0} & 407 & 1696 & \textbf{2103} & --- & 4.5 \\
  & Mem0  & 1521 & --- & 1696 & 3217 & 35\% & --- \\
  & $\text{Mem0}\textsuperscript{\textit{g}}$ & 1522 & --- & 1696 & 3218 & 35\% & --- \\
\midrule
\multirow{3}{*}{\shortstack[c]{Qwen2.5\\1.5B}}
  & \method{} & \textbf{0} & 26 & 1696 & \textbf{1722} & --- & 10.6 \\
  & Mem0  & 104 & --- & 1696 & 1800 & 4\% & --- \\
  & $\text{Mem0}\textsuperscript{\textit{g}}$ & 100 & --- & 1696 & 1796 & 4\% & --- \\
\midrule
\multirow{3}{*}{\shortstack[c]{Qwen2.5\\3B}}
  & \method{} & \textbf{0} & 62 & 1696 & \textbf{1758} & --- & 2.8 \\
  & Mem0  & 1285 & --- & 1696 & 2981 & 41\% & --- \\
  & $\text{Mem0}\textsuperscript{\textit{g}}$ & 1278 & --- & 1696 & 2974 & 41\% & --- \\
\midrule
\multirow{3}{*}{\shortstack[c]{Qwen2.5\\7B}}
  & \method{} & \textbf{0} & 193 & 1696 & \textbf{1889} & --- & 4.3 \\
  & Mem0  & 1573 & --- & 1696 & 3269 & 42\% & --- \\
  & $\text{Mem0}\textsuperscript{\textit{g}}$ & 1576 & --- & 1696 & 3272 & 42\% & --- \\
\midrule
\multirow{2}{*}{GPT-4o-mini}
  & \method{} & \textbf{0} & 601 & 1696 & \textbf{2297} & --- & 10.3 \\
  & Mem0  & 1521 & --- & 1696 & 3217 & 29\% & --- \\
\bottomrule
\end{tabular}
\vspace{0.35em}
\caption{
Complete write-side LLM-call budget on full LoCoMo (the all-backbone version of Table~\ref{tab:action-stats}), for seven backbone-matched open-weight triads plus a GPT-4o-mini \method{}-vs-Mem0 pair. \emph{Decision-stage} isolates controller cost: \method{} makes \textbf{zero routing calls} and invokes the LLM only to \emph{merge} the $\pi_{\text{upd}}$ fraction routed to \textsc{Update}, while Mem0/Mem0g fuse routing and edit into one call per non-empty \texttt{add} (Update marked ---). \emph{Total write LLM calls} adds shared fact-extraction; \emph{Total $\downarrow$} is \method{}'s reduction vs.\ that row; $\pi_{\text{upd}}$ is the empirical share of routed candidates that fall in the \textsc{Update} band. $^{\dagger}$Llama-3.2-1b is excluded from the aggregate (see text).
}
\label{tab:action-stats-all}
\end{table*}

\subsection{Full Write-Side LLM-Call Budget}
\label{app:action-stats-all}

Table~\ref{tab:action-stats} in the main text reports the write-side LLM-call budget for a representative subset of backbones. Table~\ref{tab:action-stats-all} extends this to all eight configurations: seven backbone-matched open-weight triads (\method{}, Mem0, $\text{Mem0}\textsuperscript{\textit{g}}$) plus the GPT-4o-mini \method{}-vs-Mem0 pair. Within each triad, the dataset, fact-extraction prompt, embedding model, and retrieval stack are identical; the only difference is what happens after fact extraction.

\paragraph{Decision-stage savings.}
The ``Route'' and ``Update'' columns isolate the controller cost. \method{} makes \textbf{zero routing LLM calls} on every backbone because the vMF novelty gate resolves \textsc{Add} and \textsc{Noop} in closed form; the LLM is invoked only to merge the narrow fraction $\pi_{\text{upd}}$ of candidates routed to \textsc{Update}. The empirical $\pi_{\text{upd}}$ ranges from $2.7\%$ (DeepSeek-R1-7B) to $10.6\%$ (Qwen2.5-1.5B), meaning that the vast majority of write decisions are resolved without any LLM call. In contrast, Mem0 and $\text{Mem0}\textsuperscript{\textit{g}}$ invoke a routing LLM on every non-empty \texttt{add} call, producing between $100$ and $1{,}576$ decision-stage calls depending on the backbone.

\paragraph{Total write-side reduction.}
Including the shared $1{,}696$ extraction calls (one per \texttt{add} at \texttt{batch\_size}$=8$), \method{} still reduces total write-side LLM calls by $29$--$42\%$ (mean $32\%$) on seven of the eight backbones.
The sole exception is Llama-3.2-1B (marked $^\dagger$). On this backbone, Mem0's LLM-based router emits malformed JSON on $1{,}347$ of $1{,}696$ calls ($79\%$), which artificially deflates its routing-call count: most calls are discarded as parse failures rather than counted as successful routes. Because \method{}'s closed-form gate has no such parse-failure mode, the resulting call counts are not comparable, and we exclude this backbone from the aggregate efficiency claim.

\section{Responsible Use of Artifacts}
\subsection{Artifact Use and Intended Use}
We use existing artifacts, including the LoCoMo benchmark, backbone language models, embedding models, and prior memory-system implementations, only for research and evaluation purposes in the experimental settings described in this paper. Our use is intended to be consistent with the intended use and access conditions specified by the original artifact providers, where such conditions are available. We do not claim rights over third-party artifacts, and we do not redistribute restricted datasets, proprietary model weights, or API-backed systems except as permitted by their original terms. Any artifacts released as part of this work (e.g., code, prompts, or configuration files) are intended for research use only. These released artifacts are designed to support reproducibility of the proposed method and are not intended to override or expand the original access conditions attached to the underlying third-party datasets, models, or services.
\subsection{Artifact Documentation}
Our experiments study long-term conversational memory in English using the LoCoMo evaluation protocol. We evaluate single-hop, multi-hop, temporal, and open-domain question settings, and we compare SAGE against prior memory-evolution systems under matched backbone configurations. These artifacts are used to study write-side memory control in research settings rather than to support deployment claims in real-world user-facing systems.
\end{document}